\documentclass[]{fairmeta}
\usepackage{mathpazo}
\usepackage{tgpagella}

\usepackage[table,xcdraw]{xcolor}
\usepackage{amsmath}
\usepackage{amssymb}
\usepackage{graphicx}
\usepackage{arydshln}
\usepackage{multirow}
\usepackage{booktabs}
\usepackage{marvosym}  
\usepackage{colortbl}
\usepackage{float}
\usepackage{cuted}  
\usepackage{pifont}
\usepackage{pgfplots}
\usepackage{wrapfig}
\usepackage{makecell}

\def \ie {\emph{i.e.}}
\def \eg {\emph{e.g.}}
\def \etc {\emph{etc.}}

\definecolor{darkgreen}{rgb}{0.04,0.63,0.07}
\definecolor{skyblue}{rgb}{0.04,0.40,0.80}
\definecolor{tablegray}{gray}{.9}

\title{Scaling Zero-Shot Reference-to-Video Generation}

\author[1,2,*]{Zijian Zhou}
\author[1]{Shikun Liu}
\author[1]{Haozhe Liu}
\author[1]{Haonan Qiu}
\author[1]{Zhaochong An}
\author[1]{Weiming Ren}
\author[1]{Zhiheng Liu}
\author[1]{Xiaoke Huang}
\author[1]{Kam Woh Ng}
\author[1]{Tian Xie}
\author[1]{Xiao Han}
\author[1]{Yuren Cong}
\author[1]{Hang Li}
\author[1]{Chuyan Zhu}
\author[1]{Aditya Patel}
\author[1]{Tao Xiang}
\author[1]{Sen He}

\affiliation[1]{Meta AI}
\affiliation[2]{King's College London}

\contribution[*]{Work done at Meta}

\abstract{
Reference-to-video (R2V) generation aims to synthesize videos that align with a text prompt while preserving the subject identity from reference images.
However, current R2V methods are hindered by the reliance on explicit reference image-video-text triplets, whose construction is highly expensive and difficult to scale.
We bypass this bottleneck by introducing Saber, a scalable zero-shot framework that requires no explicit R2V data.
Trained exclusively on video-text pairs, Saber employs a masked training strategy and a tailored attention-based model design to learn identity-consistent and reference-aware representations.
Mask augmentation techniques are further integrated to mitigate copy-paste artifacts common in reference-to-video generation.
Moreover, Saber demonstrates remarkable generalization capabilities across a varying number of references and achieves superior performance on the OpenS2V-Eval benchmark compared to methods trained with R2V data.

}

\date{\today}

\metadata[Website]{\url{https://franciszzj.github.io/Saber/}}

\begin{document}
\maketitle

\begin{figure}[h]
    \centering
    \vspace{-2mm}
    \includegraphics[width=\textwidth]{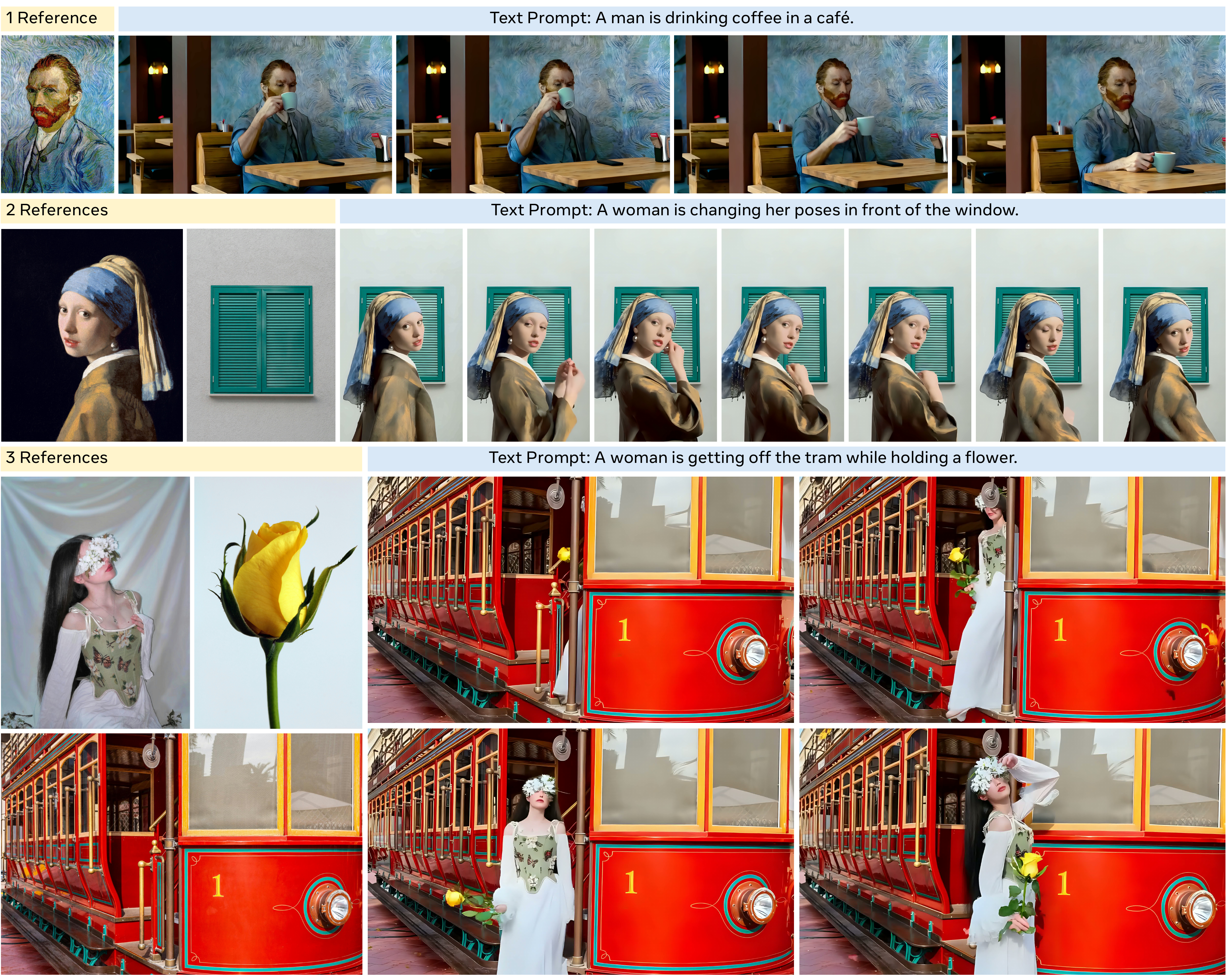}
    \vspace{-6mm}
    \caption{
    Saber is a zero-shot reference-to-video method trained only on video-text pairs.
    It preserves identity and appearance while coherently integrating single/multiple references into videos guided by text prompts.
    }
    \vspace{-2mm}
    \label{fig:teaser}
\end{figure}

\section{Introduction}
\label{sec:introduction}

Reference-to-video (R2V) generation synthesizes videos that align with a given text prompt while preserving the identity and appearance of subjects in reference images.
This task represents a crucial step toward personalized video generation, enabling applications such as customized storytelling~\citep{rahman2023make, wang2024evolving, zhou2024storydiffusion} and virtual avatars~\citep{guo2024liveportrait, yuan2025identity, gao2025identity, shen2025identity}.
Despite recent progress in text-to-video (T2V) and image-to-video (I2V) generation~\citep{hong2022cogvideo, yang2024cogvideox, liu2024mardini, hacohen2024ltx, kong2024hunyuanvideo, wan2025wan, gao2025seedance, zhang2025waver}, R2V remains uniquely challenging as it must simultaneously ensure semantic alignment with text and maintain high-fidelity subject identity from the reference images.

Existing R2V methods~\citep{zhou2024sugar, liu2025phantom, jiang2025vace, deng2025magref, fei2025skyreels, hu2025polyvivid, xue2025stand, li2025bindweave} typically rely on constructing explicit R2V datasets (\eg, OpenS2V-5M~\citep{yuan2025opens2v} and Phantom-Data~\citep{chen2025phantom}) that contain triplets of reference images, videos, and text prompts.
Building such datasets involves complex pipelines for data collection, annotation, clustering and filtering, which are costly and difficult to scale.
Moreover, the limited diversity of reference images in these datasets restrict generalization, making it difficult to handle unseen subject categories.

We propose Saber, a scalable, zero-shot framework that bypasses this data bottleneck.
Saber is trained solely on large-scale video-text pairs, the same data paradigm used for T2V and I2V models.
This design allows Saber to leverage abundant video-text datasets~\citep{chen2024panda, wang2025koala, sstk}, completely eliminating the need for bespoke R2V data construction.

To this end, our method introduces a masked training strategy that uses randomly sampled and partially masked video frames as reference images during training, where the randomness of masking provides diverse reference conditions and improves generalization across subject categories.
This process compels the model to learn identity- and appearance-consistent representations from the reference context, effectively simulating the R2V task without R2V data.
This strategy is complemented by a tailored attention mechanism, guided by attention masks, which directs the model to focus on reference-aware features while suppressing background noise.
To further enhance visual fidelity and mitigate the copy-paste artifacts which is common in reference-to-video generation~\cite{liu2025phantom, fei2025skyreels, hu2025hunyuancustom}, we integrate a series of spatial mask augmentations, effectively improving the visual quality of the generated videos.

Saber's design is inherently scalable.
It naturally supports a varying number of reference images (see Fig.~\ref{fig:teaser} and Fig.~\ref{fig:vis_opens2v_eval}), without additional data preparation or modification to the training pipeline, allowing for richer, multi-subject customization.
The stochasticity of the masked training strategy also allows Saber to robustly handle multiple reference views of the same subject (see Fig.~\ref{fig:one_subject_multi_view}).

We evaluate Saber on the OpenS2V-Eval~\citep{yuan2025opens2v} benchmark, where it consistently outperforms models~\citep{zhou2024sugar, liu2025phantom, jiang2025vace, deng2025magref, fei2025skyreels, hu2025polyvivid, li2025bindweave} that were explicitly trained on R2V data.
In addition, by simply adjusting the masking ratio during training, Saber can adapt to references depicting either foreground subjects or background scenes (see Fig.~\ref{fig:teaser}).

Our contributions are summarized as follows:
\begin{itemize}
    \item We introduce Saber, the first zero-shot R2V framework that eliminates the need for explicit R2V data through masked training on video-text pairs.
    \item Saber surpasses previous R2V-data-trained methods on OpenS2V-Eval~\citep{yuan2025opens2v} and demonstrates strong generalization and scalability, paving the way for future research in scaling reference-to-video generation.
\end{itemize}

\vspace{-2mm}
\section{Related Work}
\label{sec:related_work}

\subsection{Video Generation}
The rapid progress of diffusion models~\citep{rombach2022high} has greatly advanced video generation.
Early methods~\citep{blattmann2023stable, guo2023animatediff, chen2024gentron} extended pre-trained text-to-image models~\citep{rombach2022high, podell2023sdxl} with temporal modules to synthesize videos.
More recently, large-scale models based on Diffusion Transformer~\citep{peebles2023scalable} and trained on massive video-text datasets~\citep{chen2024panda, wang2025koala} have achieved state-of-the-art, high-fidelity video generation~\citep{yang2024cogvideox, kong2024hunyuanvideo, chen2025goku, wan2025wan, gao2025seedance, zhang2025waver}. 
Despite these advances, existing methods mainly focus on text-to-video and image-to-video tasks.
While fine-grained, subject-driven control, as required by reference-to-video generation, remains a significant challenge, the complex, costly data construction for R2V datasets~\citep{yuan2025opens2v, liu2025phantom} makes the large-scale training seen in T2V and I2V infeasible.

\subsection{Reference-to-Video Generation}
Building on the progress of text-to-video and image-to-video models~\citep{hong2022cogvideo, yang2024cogvideox, kong2024hunyuanvideo, hacohen2024ltx, wan2025wan}, reference-to-video generation~\citep{yuan2025identity, jiang2025vace, liu2025phantom, hu2025hunyuancustom, hu2025polyvivid, li2025bindweave} has seen significant advancement.
Early studies~\citep{gao2025identity, yuan2025identity, shen2025identity} mainly focused on human reference images, termed identity-preserving video generation, where facial or body features are injected into models to maintain identity consistency.
Later, reference images extended from humans to various objects and backgrounds~\citep{liu2025phantom, jiang2025vace, hu2025hunyuancustom}, allowing more flexible control.
Some works~\citep{liu2025phantom, yuan2025opens2v} also refer to this task as subject-consistent or subject-to-video generation, which is equivalent to reference-to-video generation.

Representative works include Phantom~\citep{liu2025phantom}, which learns cross-modal alignment with a joint text-image injection model using image-video-text triplet data.
VACE~\citep{jiang2025vace} introduces a context adapter to process reference images and enable temporal-spatial feature interaction within a unified framework.
SkyReels-A2~\citep{fei2025skyreels} builds an image-text joint embedding model to inject multi-element representations, balancing consistency and coherence.
HunyuanCustom~\citep{hu2025hunyuancustom} employs a LLaVA-based~\citep{liu2023visual} fusion module and an image ID enhancement module to strengthen multimodal understanding and identity consistency.
MAGREF~\citep{deng2025magref} uses region-aware masking and pixel-wise concatenation for effective multi-reference interaction.
PolyVivid~\citep{hu2025polyvivid} adds a 3D-RoPE enhancement and attention-inherited identity injection to reduce identity drift.
BindWeave~\citep{li2025bindweave} leverages an MLLM~\citep{bai2025qwen2} to link complex prompts with visual subjects, improving video generation quality.

However, a critical limitation unites these approaches: these methods rely on explicit reference image-video-text triplet datasets, which are costly and difficult to construct.
Datasets such as OpenS2V-5M~\citep{yuan2025opens2v} and Phantom-Data~\citep{chen2025phantom} require complex construction pipelines, including candidate extraction, low-quality sample filtering, sample clustering, cross-pair matching, and expensive API calls for reference image generation.
Such processes result in uncontrolled data quality, poor scalability, and high construction complexity.
In contrast, we propose a zero-shot R2V framework trained solely on video-text pairs, achieving strong performance on public benchmarks.

\section{Preliminary}
\label{sec:preliminary}

\subsection{Video Generation Models}
\label{sec:video_model}
Video generation models~\citep{hong2022cogvideo, yang2024cogvideox, kong2024hunyuanvideo, hacohen2024ltx, wan2025wan} have achieved remarkable progress and gained broad attention.
Among these, the Wan Video series (\eg, Wan2.1~\citep{wan2025wan}) is one of the most popular open-source frameworks.
Our method builds on the Wan2.1-14B model~\citep{wan2025wan}, which consists of a variational autoencoder (VAE)~\citep{kingma2013auto}, a transformer backbone~\citep{vaswani2017attention, peebles2023scalable} and a text encoder (\ie, umt5-xxl~\citep{chung2023unimax}).
The VAE encodes videos into temporally and spatially compressed latents ${\mathbf{z}_{0}}$ and decodes them back to pixel space, reducing token count and computation.
Wan2.1 trains the diffusion model $\Psi$ using Flow Matching (FM)~\citep{lipman2022flow}, where the forward process linearly interpolates between data and noise.
For a time step $t \in [0, 1]$, Gaussian noise $\epsilon \sim \mathcal{N}(0, I)$ is added to $\mathbf{z}_{0}$ to obtain $\mathbf{z}_{t}$, following $\mathbf{z}_{t} = (1 - t) \mathbf{z}_{0} + t \epsilon$.
The model is optimized to predict the target velocity with the following objective:
\begin{align}
\label{eq:loss}
    \mathcal{L}_\text{FM} = \mathbb{E}_{\mathbf{z}_{0}, \epsilon, t, c} \left[\left\|(\mathbf{z}_{0} - \epsilon)  - \Psi_\theta ( \mathbf{z}_{t}, t, c)\right\|_{2}^{2} \right],
\end{align}
where $\theta$ denotes the learnable parameters of the diffusion model $\Psi$, and $c$ represents the condition features derived from the given text prompt and reference images.

\subsection{Task Definition and Notations}
\label{sec:task_define}
Given $K$ reference images $\{ \mathbf{I}_{k} \in \mathbb{R}^{H_{k} \times W_{k} \times 3}\}_{k=1}^{K}$ and a text prompt $\mathbf{P}$, the reference-to-video method generates a video $\{ \mathbf{I}_{f} \in \mathbb{R}^{H \times W \times 3} \}_{f=1}^{F}$ whose subjects preserve the identities and appearances of those in the reference images while following the instructions in text prompt $\mathbf{P}$.
Here, $H_{k}$ and $W_{k}$ denote the height and width of the corresponding $k$-th reference image $\mathbf{I}_{k}$, while $F$, $H$, and $W$ represent the number of frames, height, and width of the generated video.

\section{Method}
\label{sec:method}

Our goal is to train a diffusion model $\Psi_\theta$ capable of generating videos $\{ \mathbf{I}_{f} \}_{f=1}^{F}$ that preserve the identity and appearance of subjects in the given reference images $\{ \mathbf{I}_{k} \}_{k=1}^{K}$ while following the provided text prompt $\mathbf{P}$. 
Previous methods~\citep{liu2025phantom, jiang2025vace, hu2025hunyuancustom, hu2025polyvivid, li2025bindweave} rely on reference image-video-text triplets, which are costly and hard to scale.
In contrast, Saber achieves R2V capabilities using only video-text pairs, the same data paradigm used for T2V and I2V training.

Our core idea of Saber is to simulate the R2V task by replacing the explicitly collected reference images with randomly masked frames during training.
This masked training strategy is supported by two key components to enhance robustness and visual quality: i) a series of mask augmentations designed to mitigate copy-paste artifacts, and ii) a tailored attention mechanism that guides the model to focus on relevant reference features.

We first introduce the construction of masked frames in Sec.~\ref{sec:masked_frames}, including mask generation and augmentation.
Next, we present the model's architecture design in Sec.~\ref{sec:model_design}, detailing the input format and transformer-based attention mechanism.
Finally, Sec.~\ref{sec:inference} describes the zero-shot R2V inference process.

\subsection{Masked Frames as Reference}
\label{sec:masked_frames}

A standard R2V model learns to extract identity and appearance features from reference images $\{ \mathbf{I}_k \}_{k=1}^{K}$ and inject them into the generated video $\{ \mathbf{I}_{f} \}_{f=1}^{F}$.
However, existing R2V datasets~\citep{yuan2025opens2v, chen2025phantom} mainly consist of humans and common objects, leading to limited subject diversity and poor generalization.
To address this, instead of relying on pre-collected reference images, we use randomly masked frames as dynamic substitutes during training. 
This strategy naturally introduces highly diverse reference samples, allowing the model to learn more effective subject integration and achieve stronger generalization.

\begin{wrapfigure}{r}{0.6\textwidth}
    \centering
    \includegraphics[width=1.0\linewidth]{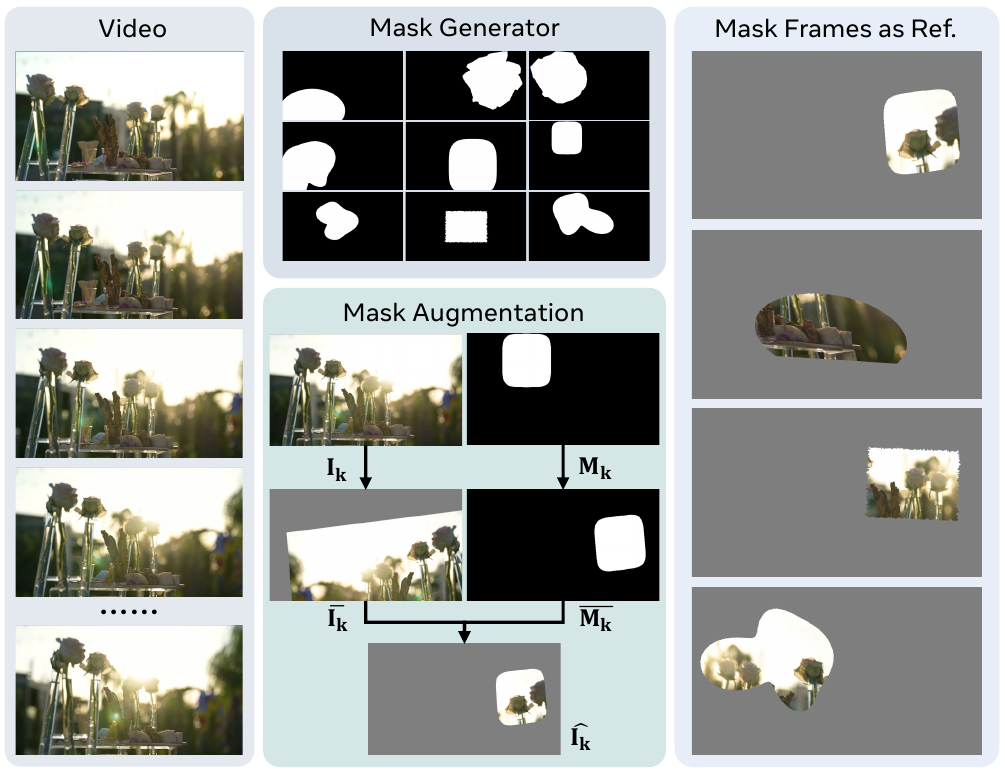}
    \vspace{-6mm}
    \caption{
    \textbf{Masked reference generation.}
    Given a video, the mask generator produces diverse random masks, which are then applied to each randomly sampled video frame with mask augmentation.
    }
    \label{fig:mask_frames}
\end{wrapfigure}

As shown in Fig.~\ref{fig:mask_frames}, for each $k$-th reference image $\mathbf{I}_k$ randomly sampled from the video, we first use a \textbf{mask generator} to produce a binary mask $\mathbf{M}_{k} \in \{0, 1\}^{H \times W}$.
To mitigate the copy-paste issue~\citep{liu2025phantom} in R2V tasks, we perform \textbf{mask augmentation} to disrupt spatial correspondence between the masked references and their corresponding video frames.
Specifically, we apply an identical set of spatial augmentations to both $\mathbf{I}_k$ and $\mathbf{M}_k$, producing $\bar{\mathbf{I}}_{k}$ and $\bar{\mathbf{M}}_{k}$.
The masked frame $\hat{\mathbf{I}_{k}}$ is then obtained as $\hat{\mathbf{I}}_{k} = \bar{\mathbf{I}}_{k} \odot \bar{\mathbf{M}}_{k}$.
This process is repeated to create the full set of $K$ masked frames $\{ \hat{\mathbf{I}}_{k} \}_{k=1}^{K}$ that serve as the reference condition.

\paragraph{Mask Generator.}
We randomly select one mask type from predefined shape categories (\eg, ellipse, Fourier blob, convex/concave polygon, \etc) to generate a binary mask $\mathbf{M} \in \{ 0, 1 \}^{H \times W}$ with a target foreground area ratio $r \in [r_\text{min}, r_\text{max}]$.
Specifically, we first randomly select a foreground center.
To ensure that the generated mask meets the desired foreground area ratio $r$, we define a continuous scale parameter for each shape category, where the mask’s foreground area increases monotonically with the scale.
A bisection search over the scale is then performed to satisfy the area ratio constraint.
When pixel discretization prevents an exact match, small topology-preserving adjustments are applied: ``growth'' dilates background boundary pixels, while ``shrinkage'' erases background boundary pixels.
This design ensures controllable foreground area ratios while maintaining diversity in mask shapes.
Several mask examples are illustrated in Fig.~\ref{fig:mask_frames} Top.

\paragraph{Mask Augmentation.}
We apply random affine transformations, including rotation, scaling, shear, translation, and optional horizontal flip, to both the image $\mathbf{I}_{k}$ and its mask $\mathbf{M}_{k}$, ensuring the masked region remains fully inside the frame.
Transformation parameters are uniformly sampled within predefined ranges and validated to avoid boundary overflow.
The same affine transformation is applied to the image and mask using bilinear and nearest-neighbor interpolation, respectively.

The reference code of the mask generator and augmentation are provided in the supplementary materials.

\subsection{Model Design}
\label{sec:model_design}

After obtaining the masked frames $\{ \hat{\mathbf{I}}_{k} \}_{k=1}^{K}$ as reference images, we detail our model design for the R2V task.
We adopt a simple yet effective \textbf{input format} by concatenating reference images along the temporal dimension at the end of the target video frames in latent space.
This allows the model to manage the interaction between the target video latents and reference latents through the \textbf{attention mechanism} in each transformer block.

\paragraph{Input Format.}
Given a video-text pair $\{ \mathbf{I}_{f} \}_{f=1}^{F}$ and $\mathbf{P}$, and the masked frames $\{ \hat{\mathbf{I}}_{k} \}_{k=1}^{K}$ as reference images, we use the VAE to encode the video from pixel space into latent space, obtaining $\mathbf{z}_{0} = \{ \mathbf{z}_{\hat{f}} \in \mathbb{R}^{h \times w \times d} \}_{\hat{f}=1}^{\hat{F}}$.
Here, $\hat{F} = \lfloor(F - 1) / 4\rfloor + 1$, where $4$ is the temporal compression ratio of the Wan2.1 VAE~\citep{wan2025wan}, $h$, $w$, and $d$ denote the height, width, and feature dimension of the video latent, respectively.
We obtain $\mathbf{z}_{t}$ with time step $t$ following Sec.~\ref{sec:video_model}.
For the reference images, we individually encode each $\hat{\mathbf{I}}_{k}$ using the VAE to obtain $\mathbf{z}_\text{ref} = \{ \mathbf{z}_{k} \in \mathbb{R}^{h \times w \times d} \}_{k=1}^{K}$.
Accordingly, each $\mathbf{M}_{k}$ is resized to match the latent space resolution, producing $\mathbf{m}_\text{ref} = \{ \mathbf{m}_{k} \in \{0,1\}^{h \times w \times 4} \}_{k=1}^{K}$, where $0$ indicates non-reference and $1$ indicates reference region.
The transformer input $\mathbf{z}_\text{in}$ is defined as in Eq.~\ref{eq:input_format}:
\begin{equation}
\label{eq:input_format}
\mathbf{z}_{\text{in}} = \mathsf{cat}
\begin{bmatrix}
    \mathsf{cat}[ & \mathbf{z}_{t}         & \mathbf{z}_\text{ref} & ]_\text{temporal} \\
    \mathsf{cat}[ & \mathbf{m}_\text{zero} & \mathbf{m}_\text{ref} & ]_\text{temporal} \\
    \mathsf{cat}[ & \mathbf{z}_\text{zero} & \mathbf{z}_\text{ref} & ]_\text{temporal}
\end{bmatrix}_\text{channel},
\end{equation}
where $\mathsf{cat}[\cdot]_\text{temporal}$ and $\mathsf{cat}[\cdot]_\text{channel}$ denote concatenation along the temporal and channel dimensions, respectively. $\mathbf{z}_\text{zero}$ is the VAE-encoded latent of a zero-value video, and $\mathbf{m}_\text{zero}$ is an all-zero mask, both shaped to match the temporal dimensions of the video part.
Note that $\mathbf{z}_\text{ref}$ remains noise-free to preserve accurate conditioning.

\begin{wrapfigure}{r}{0.6\textwidth}
    \centering
    \includegraphics[width=1.0\linewidth]{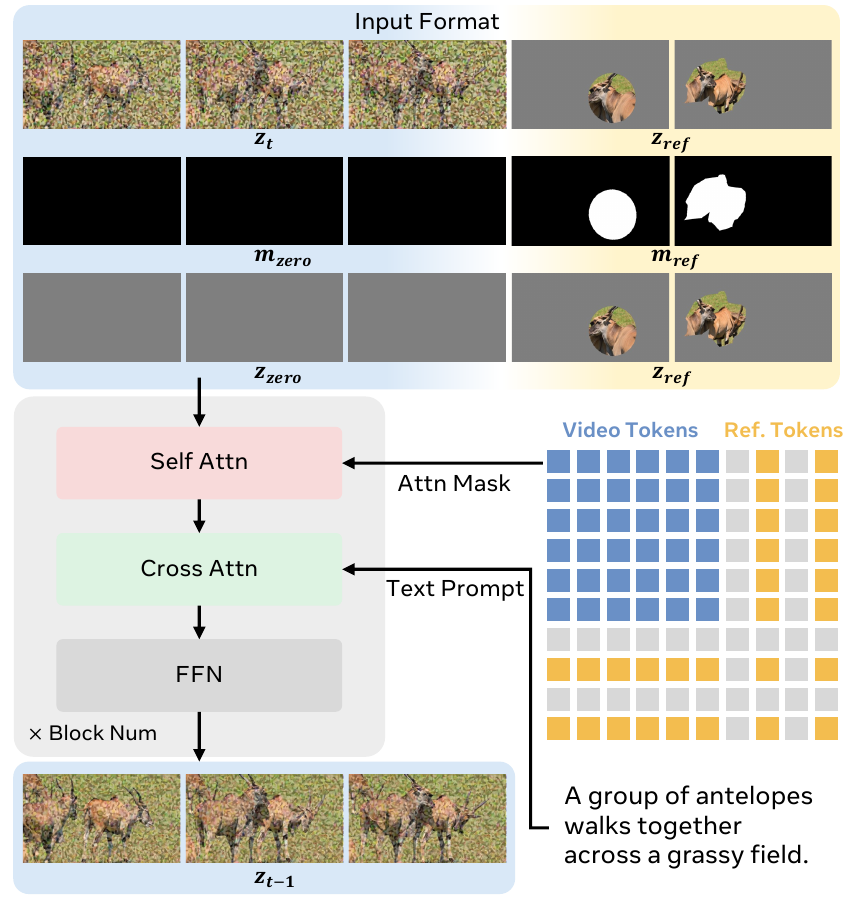}
    \vspace{-6mm}
    \caption{
    \textbf{Model design overview.}
    Masked frames serve as reference images and are concatenated to the video tokens in latent space.
    Self-attention enables interaction between video and reference tokens under the attention mask, while cross-attention incorporates text guidance for semantic alignment.
    The VAE, text encoder, and timestep components are omitted for clarity.
    }
    \label{fig:model_design}
    \vspace{-4mm}
\end{wrapfigure}

\paragraph{Attention Mechanism.}
After obtaining the transformer input $\mathbf{z}_\text{in}$, we encode the text prompt $\mathbf{P}$ into text features $\mathbf{z}_\text{P}$ and jointly feed them, along with the time step $t$, into the transformer.
Each transformer block consists of a self-attention, cross-attention, and feed-forward (FFN) modules.

In self-attention, the video and reference parts of $\mathbf{z}_\text{in}$ interact with each other.
To avoid attending to non-reference regions, each $\mathbf{M}_{k}$ is resized to match the flattened latent shape to form an attention mask, where video tokens are bi-directionally attended, and only valid reference regions are attended in the reference part.

The self-attention output is then passed to the cross-attention module to interact with $\mathbf{z}_\text{P}$.
Here, video tokens are guided by the text prompt, while reference tokens learn their semantic alignment, enabling the integration of reference image information under textual constraints.
The FFN module refines the results, with the time step $t$ injected into the latents for the control of the time step.
After multiple transformer blocks, the transformer model outputs the predicted latent $\mathbf{z}_{t-1}$.
The model design is shown in Fig.~\ref{fig:model_design}.

\subsection{Zero-Shot Inference}
\label{sec:inference}

In this section, we present the approach that enables the model trained with masked frames to perform zero-shot R2V inference.
During inference, for each reference image $\mathbf{I}_{k}$, we first use a pre-trained object segmenter~\citep{ren2024grounded, zheng2024bilateral} to extract the foreground subject region mask $\mathbf{M}_{k}$.
We then normalize the reference image $\mathbf{I}_{k}$ to the range of $[-1, 1]$ and fill the masked background regions with zeros (color gray).
Notably, this segmentation step is flexible.
If a reference image is intended to provide a background scene rather than a foreground subject, segmentation is skipped.
In this case, we use the full, unmasked reference image and an all-ones mask $\mathbf{M}_{k}$, treating the entire image as the reference region.

Both $\mathbf{I}_{k}$ and $\mathbf{M}_{k}$ are processed by a resize-and-padding operation, which scales $\mathbf{I}_{k}$ from its original size $(H_{k}, W_{k})$ to fit within the target video size $(H, W)$ while preserving the aspect ratio and pads the remaining area with zeros to produce a centered reference image of size $(H, W)$.
Finally, the processed reference image and mask are fed into the model following the input format in Sec.~\ref{sec:model_design} and are used for prediction following the Wan inference pipeline~\citep{wan2025wan}.

\section{Experiments}
\label{sec:experiments}

\begin{table*}[!t]
\centering
\caption{
\textbf{Quantitative results on the OpenS2V-Eval~\citep{yuan2025opens2v} benchmark.}
Saber outperforms both closed-source and explicitly trained R2V methods, achieving the highest overall score in a zero-shot setting.
It also attains the best NexusScore for subject consistency and competitive performance on GmeScore and NaturalScore.
}
\vspace{-2mm}
\resizebox{\linewidth}{!}{
\begin{tabular}{lcccccccc}
\toprule[1.5pt]
\textbf{Method} & \textbf{Total Score} $\uparrow$ & \textbf{Aesthetics} $\uparrow$ & \textbf{MotionSmoothness} $\uparrow$ & \textbf{MotionAmplitude} $\uparrow$ & \textbf{FaceSim} $\uparrow$ & \textbf{GmeScore} $\uparrow$ & \textbf{NexusScore} $\uparrow$ & \textbf{NaturalScore} $\uparrow$ \\
\midrule
\multicolumn{9}{c}{\cellcolor[HTML]{EFEFEF}\textit{Closed-source commercial R2V methods}} \\
Pika2.1~\citep{pika}                           & 51.88\%  & 46.88\% & 87.06\% & 24.71\% & 30.38\% & 69.19\% & 45.40\% & 63.32\% \\
Vidu2.0~\citep{vidu}                           & 51.95\%  & 41.48\% & 90.45\% & 13.52\% & 35.11\% & 67.57\% & 43.37\% & 65.88\% \\
Kling1.6~\citep{kling}                         & 56.23\%  & 44.59\% & 86.93\% & 41.60\% & 40.10\% & 66.20\% & 45.89\% & 74.59\% \\
\multicolumn{9}{c}{\cellcolor[HTML]{EFEFEF}\textit{Explicit R2V data-based training methods}} \\
SkyReels-A2~\citep{fei2025skyreels}            & 52.25\%  & 39.41\% & 87.93\% & 25.60\% & 45.95\% & 64.54\% & 43.75\% & 60.32\% \\
MAGREF~\citep{deng2025magref}                  & 52.51\%  & 45.02\% & 93.17\% & 21.81\% & 30.83\% & 70.47\% & 43.04\% & 66.90\% \\
Phantom-14B~\citep{liu2025phantom}             & 56.77\%  & 46.39\% & 96.31\% & 33.42\% & 51.46\% & 70.65\% & 37.43\% & 69.35\% \\
VACE-14B~\citep{jiang2025vace}                 & 57.55\%  & 47.21\% & 94.97\% & 15.02\% & 55.09\% & 67.27\% & 44.08\% & 67.04\% \\
BindWeave~\citep{li2025bindweave}              & 57.61\%  & 45.55\% & 95.90\% & 13.91\% & 53.71\% & 67.79\% & 46.84\% & 66.85\% \\
\multicolumn{9}{c}{\cellcolor[HTML]{EFEFEF}\textit{Zero-shot R2V methods}} \\
\textbf{Saber (Ours)}                          & \cellcolor[HTML]{CFE8D9}57.91\%  & 42.42\% & 96.12\% & 21.12\% & 49.89\% & 67.50\% & \cellcolor[HTML]{CFE8D9}47.22\% & 72.55\% \\
\bottomrule[1.5pt]
\end{tabular}
}
\label{tab:opens2v_eval}
\end{table*}

\subsection{Datasets, Metrics and Implementation Details}

\paragraph{Datasets.}
Benefiting from the masked training strategy, Saber is trained exclusively on video-text pair datasets, enabling the use of data from T2V and I2V sources.
Specifically, we employ the ShutterStock Video~\citep{sstk} dataset and generate captions for all video clips using Qwen2.5-VL-Instruct~\citep{bai2025qwen2}, thus constructing the corresponding video-text pairs for training.

\paragraph{Metrics.}
To ensure fair comparison, we adopt the OpenS2V-Eval~\citep{yuan2025opens2v} benchmark and follow its official protocol for fine-grained evaluation of reference-to-video generation.
The benchmark contains 180 prompts across seven categories, spanning single-reference (face, human, entity) and multi-reference (multi-face, multi-human, human-entity) scenarios.
We report automated metrics where higher scores indicate better performance, including Aesthetics for visual quality, MotionSmoothness for temporal coherence, MotionAmplitude for motion magnitude, and FaceSim for identity preservation.
In addition, we use three OpenS2V-Eval metrics, NexusScore, NaturalScore, and GmeScore, which measure subject consistency, naturalness, and text-video alignment, respectively.

\paragraph{Implementation Details.}
Saber is finetuned from the Wan2.1-14B~\citep{wan2025wan} model using our proposed masked training strategy on video-text pair datasets.
For the mask generator, we adopt a probabilistic sampling strategy for the foreground area ratio $r$: with 10\% probability, we set $r \in [0, 0.1]$ to simulate minimal and no reference information, enabling the model to handle varying numbers of reference images; with 80\% probability, we set $r \in [0.1,0.5]$ to represent typical primary subjects; and with the remaining 10\% probability, we set $r \in [0.5,1.0]$ to help the model learn from large reference images or background scenes.
For mask augmentation, we randomly apply rotation within $[-10, 10]$ degree, scaling in the range $[0.8, 2.0]$, horizontal flipping with 50\% probability, and shearing within $[-10, 10]$ degree.
We found these augmentations to be empirically effective at overcoming copy-paste artifacts.
We train our model with the objective defined in Eq.~\ref{eq:loss}, using the AdamW optimizer with $1e^{-5}$ learning rate and a global batch size of $64$.
During inference, we use BiRefNet~\citep{zheng2024bilateral} to segment the foreground subjects from the reference images.
Following the standard setting of Wan2.1~\citep{wan2025wan}, we generate videos with $50$ denoising steps and a CFG~\citep{ho2022classifier} guidance scale of $5.0$.

\subsection{Quantitative Results}

We follow \citet{yuan2025opens2v} and conduct a comprehensive evaluation on OpenS2V-Eval~\citep{yuan2025opens2v} benchmark, with the results presented in Tab.~\ref{tab:opens2v_eval}.
The table compares three types of methods: closed-source commercial R2V methods, explicitly trained R2V methods, and our zero-shot R2V approach.
Compared with the closed-source commercial method Kling1.6~\citep{kling}, our model achieves a 1.68\% higher total score.
Among methods trained on explicit R2V datasets, our method surpasses Phantom~\citep{liu2025phantom} by 1.14\%, VACE~\citep{jiang2025vace} by 0.36\%, and BindWeave~\citep{li2025bindweave} by 0.30\%.
While these methods rely on costly explicit R2V datasets that are difficult to scale, our approach uses only text-video pairs with a masked training strategy, achieving the best overall performance in a zero-shot setting.

Among all sub-metrics, NexusScore best represents R2V performance by measuring subject consistency.
Saber achieves the highest NexusScore, exceeding Phantom by 9.79\%, VACE by 3.14\%, and BindWeave by 0.36\%.
This shows that the masked training strategy effectively learns subject features from video-text pairs in a zero-shot setting, outperforming all R2V-data-based models.
Our method also achieves competitive results on GmeScore (text-video alignment) and NaturalScore (video naturalness).

\begin{figure*}[t]
    \centering
    \includegraphics[width=1.0\linewidth]{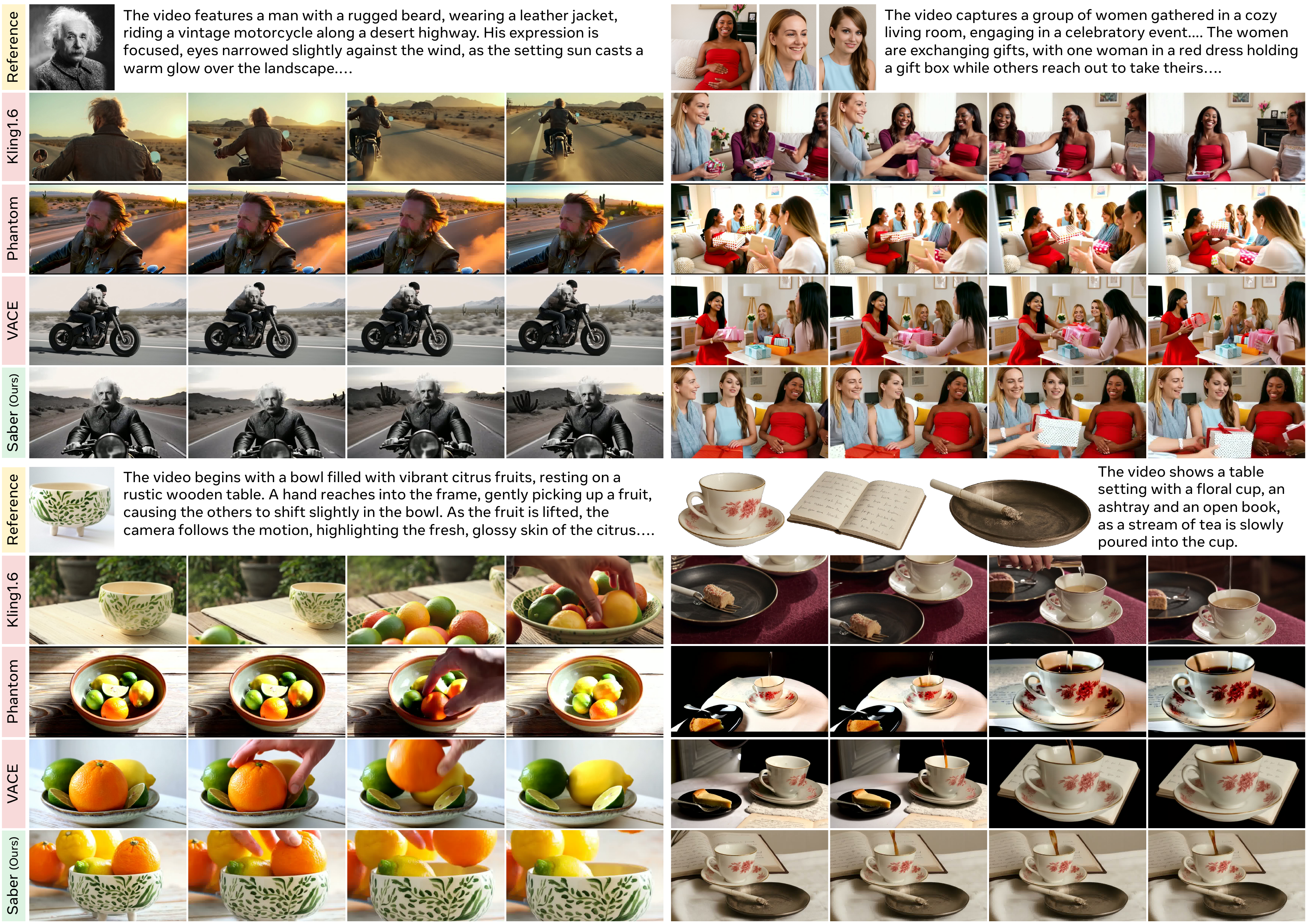}
    \vspace{-6mm}
    \caption{
    \textbf{Qualitative comparison with existing R2V methods.}
    We compare Saber with Kling1.6~\citep{kling}, Phantom~\citep{liu2025phantom}, and VACE~\citep{jiang2025vace} across four scenarios: single/multiple human and object references.
    Saber accurately preserves subject identity and appearance, integrates multiple references coherently, and generates smoother, more visually consistent videos.
    }
    \label{fig:vis_opens2v_eval}
\end{figure*}

\subsection{Qualitative Results}

We further conduct a qualitative comparison between Saber and other methods (Kling1.6~\citep{kling}, Phantom~\citep{liu2025phantom} and VACE~\citep{jiang2025vace}) across various visual scenarios, as shown in Fig.~\ref{fig:vis_opens2v_eval}.
i) In the top-left (single human reference), both Kling1.6 and Phantom fail to embed the reference subject into the generated video, leading to inconsistent facial appearances. VACE suffers from a copy-paste issue, directly overlaying the face from the reference image. In contrast, Saber generates a video with a consistent and text-aligned facial identity.
ii) In the bottom-left (single object reference), Kling1.6 produces a bowl with an incorrect leg structure, while Phantom and VACE fail to capture both the shape and appearance of the bowl from the reference. In contrast, our method, benefiting from the rich diversity of masked frames during masked training, accurately integrates the bowl’s shape and appearance into the generated video.
iii) In the top-right (multiple human references), Kling1.6 embeds only one subject, Phantom duplicates the same identity twice, and VACE fails to inject facial information from the references. In contrast, Saber incorporates all three subjects and generates a coherent, natural video.
iv) In the bottom-right (multiple object references), Kling1.6 produces incorrect patterns on the cup, while Phantom and VACE generate correct cup textures but omit the ashtray, with VACE also showing temporal discontinuity. Saber, on the other hand, generates all referenced objects correctly and maintains smooth, consistent video quality.


\subsection{Ablation Study}
We conduct a series of ablation studies to analyze the key components of Saber, including the masked training strategy, mask generator, mask augmentation, and attention mask in the attention mechanism.

\begin{table}[!t]
\centering
\caption{
\textbf{Ablation study.}
i) Masked training outperforms training on the OpenS2V-5M~\citep{yuan2025opens2v} (w/o masked training), demonstrating the advantage of the masked training strategy.
ii) Using only a single mask type reduces total score, while combining all types performs best, showing the importance of mask diversity.
iii) Fixing the foreground area ratio ($r=0.3$) leads to a further drop, indicating limited mask variation harms generalization.
}
\vspace{-2mm}
\resizebox{0.56\linewidth}{!}{
\begin{tabular}{lcccccccc}
\toprule[1.5pt]
\textbf{Method} & \textbf{Total Score} $\uparrow$ & \textbf{GmeScore} $\uparrow$ & \textbf{NexusScore} $\uparrow$ & \textbf{NaturalScore} $\uparrow$ \\
\midrule
\textbf{Saber}                                 & 57.91\%  & 67.50\% & 47.22\% & 72.55\% \\
~~~w/o masked training                         & 56.24\%  & 67.27\% & 45.33\% & 70.19\% \\
~~~ellipse only                                & 54.56\%  & 67.98\% & 40.28\% & 72.54\% \\
~~~fourier only                                & 56.33\%  & 67.25\% & 44.82\% & 72.46\% \\
~~~polygon only                                & 56.49\%  & 67.41\% & 45.24\% & 72.21\% \\
~~~fixed $r=0.3$                               & 51.73\%  & 67.12\% & 39.20\% & 69.55\% \\
\bottomrule[1.5pt]
\end{tabular}
}
\label{tab:ablation_study}
\end{table}

\paragraph{The Effect of Masked Training.}
To evaluate the effect of masked training, we finetune our model on the OpenS2V-5M~\citep{yuan2025opens2v} dataset using the same architecture.
As shown in Tab.~\ref{tab:ablation_study}, masked training improves the total score by +1.67\%, indicating that it strengthens subject representation learning and reduces overfitting to specific reference cues.

\paragraph{The Effect of Mask Generator.}
In Tab.~\ref{tab:ablation_study}, we first analyze the mask type by training the model using only one type at a time.
Using only ellipse, Fourier, or polygon masks reduces total score by 3.35\%, 1.58\%, and 1.42\%, respectively, whereas combining all types yields the best results, showing that mask diversity is crucial for masked training.
We also fix the foreground area ratio $r$ to 0.3, which leads to a 6.18\% drop, indicating that restricting mask variation limits generalization.

\begin{wrapfigure}{r}{0.5\textwidth}
    \centering
    \includegraphics[width=1.0\linewidth]{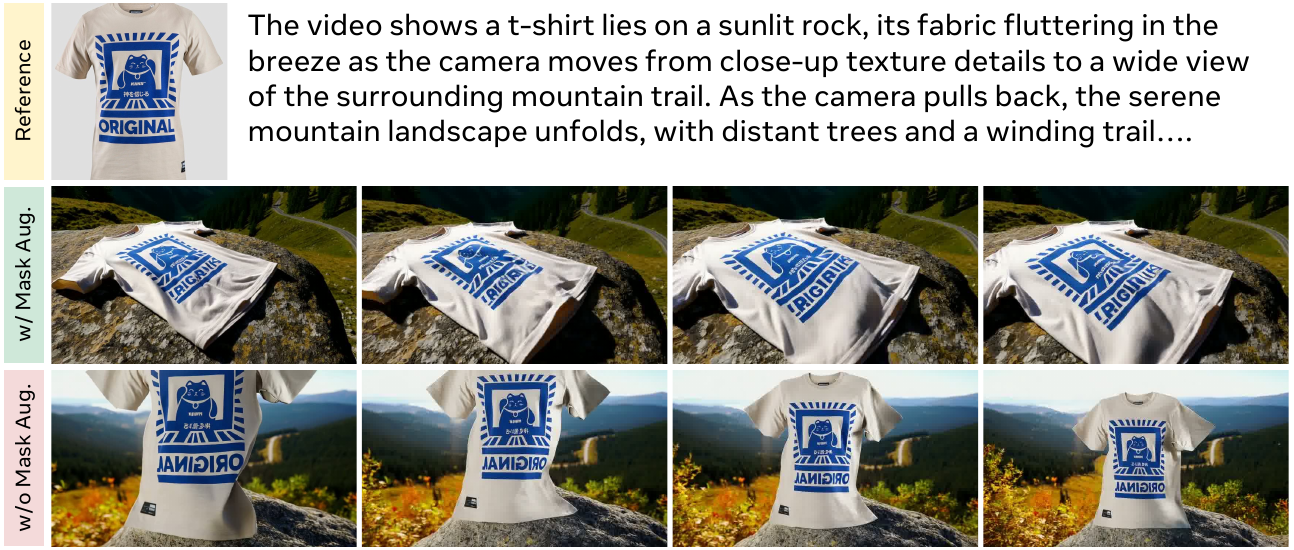}
    \vspace{-6mm}
    \caption{
    \textbf{Effect of mask augmentation.}
    Without mask augmentation, the model shows copy-paste artifacts by directly copying reference content.
    Applying augmentation enables more natural and coherent video generation.
    }
    \label{fig:ablation_mask_aug}
    \vspace{-4mm}
\end{wrapfigure}

\paragraph{The Effect of Mask Augmentation.}
We evaluate the effect of mask augmentation by training the model with and without it. As shown in Fig.~\ref{fig:ablation_mask_aug}, without augmentation, the model exhibits severe copy-paste artifacts, directly placing the T-shirt upright on the rock.
With augmentation (rotation, scaling, flipping, and shearing), the T-shirt naturally lies on the rock surface, resulting in more realistic and coherent compositions.
This demonstrates that geometric diversity in masking is crucial for natural video generation.

\begin{figure}[h]
    \centering
    \includegraphics[width=1.0\linewidth]{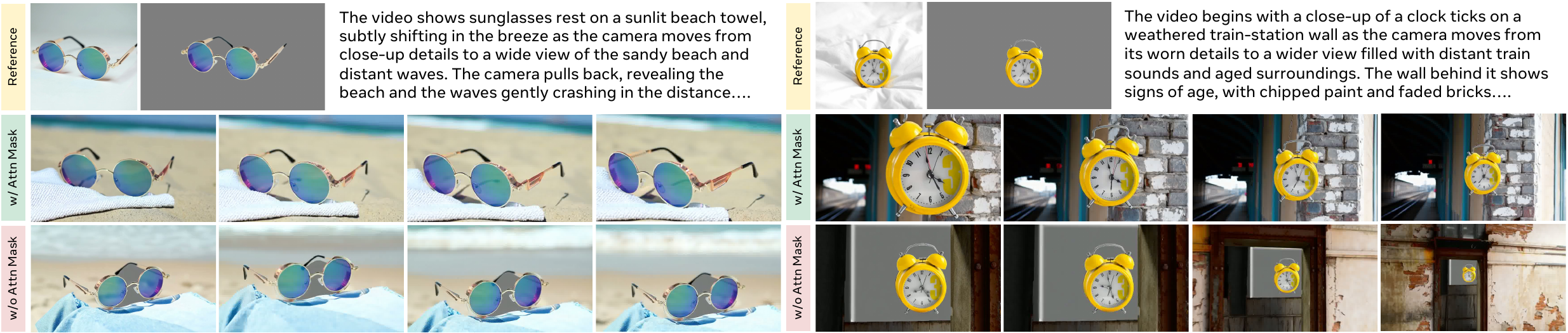}
    \vspace{-6mm}
    \caption{
    \textbf{Effect of the attention mask.}
    Removing the attention mask introduces gray artifacts around subjects, while applying it ensures clean separation from the gray background and smoother, more natural video results.
    }
    \label{fig:ablation_attn_mask}
\end{figure}

\paragraph{The Effect of the Attention Mask.}
Finally, we examine the role of the attention mask in our attention mechanism, which constrains reference-video token interaction.
As shown in Fig.~\ref{fig:ablation_attn_mask}, removing the attention mask introduces visible gray artifacts around the subject regions, as the model fails to correctly extract subjects from masked reference images (\eg, gray areas behind sunglasses or clocks). 
Incorporating the attention mask effectively resolves these issues, leading to cleaner subject separation, smoother blending, and improved overall video quality.

\subsection{Emergent Abilities}
In this section, we explore several interesting capabilities of Saber that emerged from its training strategy, demonstrating robustness beyond the standard R2V task.


\begin{wrapfigure}{r}{0.5\textwidth}
    \centering
    \includegraphics[width=1.0\linewidth]{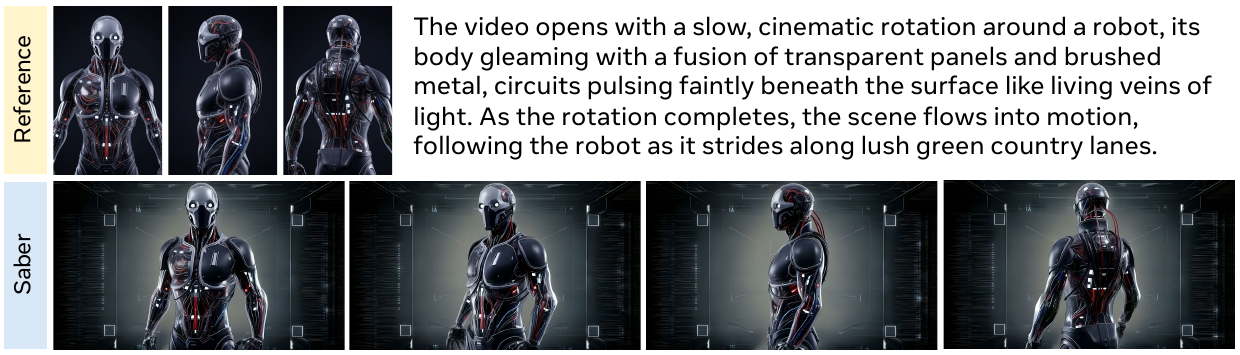}
    \vspace{-6mm}
    \caption{
    \textbf{Qualitative results on multiple reference images of the same subject.}
    Given the front, side, and back views of a robot as reference, Saber correctly recognizes them as the same subject and integrates multi-view appearance features into a coherent video, accurately preserving fine structural and surface details.
    }
    \label{fig:one_subject_multi_view}
    \vspace{-4mm}
\end{wrapfigure}

\paragraph{Single Subject Multiple Views.}
We test Saber's ability to handle multiple reference images corresponding to different views of the same subject.
As shown in Fig.~\ref{fig:one_subject_multi_view}, we use the front, side, and back views of a robot as reference inputs to Saber.
The results show that Saber successfully understands that all reference images depict the same subject (the robot) and integrates the appearance feature from different views into a single coherent video subject.
Despite the robot’s complex surface details and wiring structure, Saber accurately captures and synthesizes these visual characteristics into the generated video.

\begin{figure}[h]
    \centering
    \includegraphics[width=1.0\linewidth]{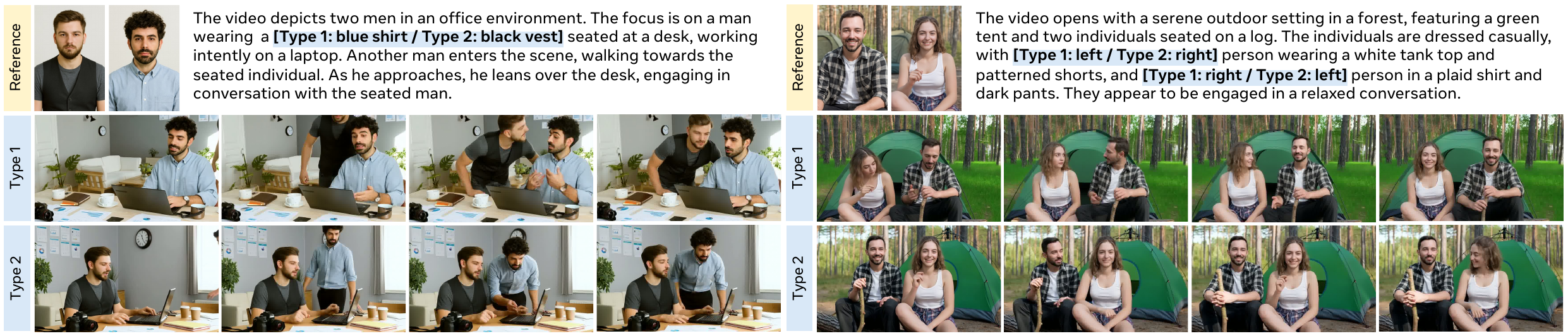}
    \vspace{-6mm}
    \caption{
    \textbf{Qualitative results of cross-modal alignment between reference images and text prompts.}
    By swapping subject descriptions in the prompts (\eg, clothing color or subject positions), Saber accurately reflects the corresponding visual changes, demonstrating robust alignment between reference images and textual descriptions through its attention mechanisms.
    }
    \label{fig:cross_modal_alignment}
    \vspace{-4mm}
\end{figure}

\paragraph{Cross-modal Alignment.}
We also evaluate the model's alignment between reference images and text prompts by swapping subject descriptions and observing the corresponding video changes.
As shown in Fig.~\ref{fig:cross_modal_alignment}, when altering prompts such as ``a man wearing a blue shirt seated'' (Type~1) and ``a man wearing a black vest seated'' (Type~2), Saber correctly aligns subjects with their descriptions.
Similarly, when swapping the positions of a man and woman in the second case, the model accurately reflects the change.
This demonstrates that, as described in Sec.~\ref{sec:model_design}, the interaction between video and reference tokens through self-attention, followed by cross-attention with text prompt features, enables robust reference image-text alignment.


\section{Conclusion}
\label{sec:conclusion}

In this work, we present Saber, a scalable zero-shot framework for reference-to-video generation that eliminates the need for explicitly R2V datasets.
Trained solely on large-scale video-text pairs, Saber leverages a masked training strategy, a tailored attention mechanism, and mask augmentation to achieve identity-consistent, natural, and coherent video generation.
It further scales to multiple references, supporting both multi-identity and multi-view inputs without additional data preparation or changes to the training pipeline.
Extensive experiments on the OpenS2V-Eval benchmark demonstrate that Saber consistently outperforms methods trained on explicit R2V data.
These results show that effective R2V models can be trained without dedicated datasets, paving the way for future research in scalable and generalizable reference-to-video generation.

\paragraph{Limitations.}
While Saber achieves strong zero-shot performance and scalability, several limitations remain.
First, R2V generation may collapse when the number of reference images increases significantly (\eg, $12$), resulting in fragmented compositions where references are combined without coherent understanding.
Second, Saber primarily focuses on identity preservation and visual coherence, while fine-grained motion control and temporal consistency under complex prompts remain challenging.
Future work can explore more effective integration of numerous reference images into unified video generation, as well as adaptive guidance to further improve controllability and realism in reference-to-video generation.

\bibliographystyle{assets/plainnat}
\bibliography{main}

\begin{thebibliography}{48}
\providecommand{\natexlab}[1]{#1}
\providecommand{\url}[1]{\texttt{#1}}
\expandafter\ifx\csname urlstyle\endcsname\relax
  \providecommand{\doi}[1]{doi: #1}\else
  \providecommand{\doi}{doi: \begingroup \urlstyle{rm}\Url}\fi

\bibitem[Bai et~al.(2025)Bai, Chen, Liu, Wang, Ge, Song, Dang, Wang, Wang, Tang, et~al.]{bai2025qwen2}
Shuai Bai, Keqin Chen, Xuejing Liu, Jialin Wang, Wenbin Ge, Sibo Song, Kai Dang, Peng Wang, Shijie Wang, Jun Tang, et~al.
\newblock Qwen2. 5-vl technical report.
\newblock \emph{arXiv preprint arXiv:2502.13923}, 2025.

\bibitem[Blattmann et~al.(2023)Blattmann, Dockhorn, Kulal, Mendelevitch, Kilian, Lorenz, Levi, English, Voleti, Letts, et~al.]{blattmann2023stable}
Andreas Blattmann, Tim Dockhorn, Sumith Kulal, Daniel Mendelevitch, Maciej Kilian, Dominik Lorenz, Yam Levi, Zion English, Vikram Voleti, Adam Letts, et~al.
\newblock Stable video diffusion: Scaling latent video diffusion models to large datasets.
\newblock \emph{arXiv preprint arXiv:2311.15127}, 2023.

\bibitem[Chen et~al.(2024{\natexlab{a}})Chen, Xu, Ren, Cong, He, Xie, Sinha, Luo, Xiang, and Perez-Rua]{chen2024gentron}
Shoufa Chen, Mengmeng Xu, Jiawei Ren, Yuren Cong, Sen He, Yanping Xie, Animesh Sinha, Ping Luo, Tao Xiang, and Juan-Manuel Perez-Rua.
\newblock Gentron: Diffusion transformers for image and video generation.
\newblock In \emph{CVPR}, 2024{\natexlab{a}}.

\bibitem[Chen et~al.(2025{\natexlab{a}})Chen, Ge, Zhang, Zhang, Zhu, Yang, Hao, Wu, Lai, Hu, et~al.]{chen2025goku}
Shoufa Chen, Chongjian Ge, Yuqi Zhang, Yida Zhang, Fengda Zhu, Hao Yang, Hongxiang Hao, Hui Wu, Zhichao Lai, Yifei Hu, et~al.
\newblock Goku: Flow based video generative foundation models.
\newblock In \emph{CVPR}, 2025{\natexlab{a}}.

\bibitem[Chen et~al.(2024{\natexlab{b}})Chen, Siarohin, Menapace, Deyneka, Chao, Jeon, Fang, Lee, Ren, Yang, et~al.]{chen2024panda}
Tsai-Shien Chen, Aliaksandr Siarohin, Willi Menapace, Ekaterina Deyneka, Hsiang-wei Chao, Byung~Eun Jeon, Yuwei Fang, Hsin-Ying Lee, Jian Ren, Ming-Hsuan Yang, et~al.
\newblock Panda-70m: Captioning 70m videos with multiple cross-modality teachers.
\newblock In \emph{CVPR}, 2024{\natexlab{b}}.

\bibitem[Chen et~al.(2025{\natexlab{b}})Chen, Li, Ma, Liu, Liu, Zhang, Li, Li, Zhou, He, et~al.]{chen2025phantom}
Zhuowei Chen, Bingchuan Li, Tianxiang Ma, Lijie Liu, Mingcong Liu, Yi~Zhang, Gen Li, Xinghui Li, Siyu Zhou, Qian He, et~al.
\newblock Phantom-data: Towards a general subject-consistent video generation dataset.
\newblock \emph{arXiv preprint arXiv:2506.18851}, 2025{\natexlab{b}}.

\bibitem[Chung et~al.(2023)Chung, Constant, Garcia, Roberts, Tay, Narang, and Firat]{chung2023unimax}
Hyung~Won Chung, Noah Constant, Xavier Garcia, Adam Roberts, Yi~Tay, Sharan Narang, and Orhan Firat.
\newblock Unimax: Fairer and more effective language sampling for large-scale multilingual pretraining.
\newblock \emph{arXiv preprint arXiv:2304.09151}, 2023.

\bibitem[Deng et~al.(2025)Deng, Guo, Yin, Fang, Yang, Wang, Yuan, Wang, Liu, Huang, et~al.]{deng2025magref}
Yufan Deng, Xun Guo, Yuanyang Yin, Jacob~Zhiyuan Fang, Yiding Yang, Yizhi Wang, Shenghai Yuan, Angtian Wang, Bo~Liu, Haibin Huang, et~al.
\newblock Magref: Masked guidance for any-reference video generation.
\newblock \emph{arXiv preprint arXiv:2505.23742}, 2025.

\bibitem[Fei et~al.(2025)Fei, Li, Qiu, Wang, Dou, Wang, Xu, Fan, Chen, Li, et~al.]{fei2025skyreels}
Zhengcong Fei, Debang Li, Di~Qiu, Jiahua Wang, Yikun Dou, Rui Wang, Jingtao Xu, Mingyuan Fan, Guibin Chen, Yang Li, et~al.
\newblock Skyreels-a2: Compose anything in video diffusion transformers.
\newblock \emph{arXiv preprint arXiv:2504.02436}, 2025.

\bibitem[Gao et~al.(2025{\natexlab{a}})Gao, Hua, Chen, Peng, and Liu]{gao2025identity}
Jiayi Gao, Changcheng Hua, Qingchao Chen, Yuxin Peng, and Yang Liu.
\newblock Identity-preserving text-to-video generation via training-free prompt, image, and guidance enhancement.
\newblock \emph{arXiv preprint arXiv:2509.01362}, 2025{\natexlab{a}}.

\bibitem[Gao et~al.(2025{\natexlab{b}})Gao, Guo, Hoang, Huang, Jiang, Kong, Li, Li, Li, Li, et~al.]{gao2025seedance}
Yu~Gao, Haoyuan Guo, Tuyen Hoang, Weilin Huang, Lu~Jiang, Fangyuan Kong, Huixia Li, Jiashi Li, Liang Li, Xiaojie Li, et~al.
\newblock Seedance 1.0: Exploring the boundaries of video generation models.
\newblock \emph{arXiv preprint arXiv:2506.09113}, 2025{\natexlab{b}}.

\bibitem[Guo et~al.(2024{\natexlab{a}})Guo, Zhang, Liu, Zhong, Zhang, Wan, and Zhang]{guo2024liveportrait}
Jianzhu Guo, Dingyun Zhang, Xiaoqiang Liu, Zhizhou Zhong, Yuan Zhang, Pengfei Wan, and Di~Zhang.
\newblock Liveportrait: Efficient portrait animation with stitching and retargeting control.
\newblock \emph{arXiv preprint arXiv:2407.03168}, 2024{\natexlab{a}}.

\bibitem[Guo et~al.(2024{\natexlab{b}})Guo, Yang, Rao, Liang, Wang, Qiao, Agrawala, Lin, and Dai]{guo2023animatediff}
Yuwei Guo, Ceyuan Yang, Anyi Rao, Zhengyang Liang, Yaohui Wang, Yu~Qiao, Maneesh Agrawala, Dahua Lin, and Bo~Dai.
\newblock Animatediff: Animate your personalized text-to-image diffusion models without specific tuning.
\newblock In \emph{ICLR}, 2024{\natexlab{b}}.

\bibitem[HaCohen et~al.(2024)HaCohen, Chiprut, Brazowski, Shalem, Moshe, Richardson, Levin, Shiran, Zabari, Gordon, et~al.]{hacohen2024ltx}
Yoav HaCohen, Nisan Chiprut, Benny Brazowski, Daniel Shalem, Dudu Moshe, Eitan Richardson, Eran Levin, Guy Shiran, Nir Zabari, Ori Gordon, et~al.
\newblock Ltx-video: Realtime video latent diffusion.
\newblock \emph{arXiv preprint arXiv:2501.00103}, 2024.

\bibitem[Ho and Salimans(2022)]{ho2022classifier}
Jonathan Ho and Tim Salimans.
\newblock Classifier-free diffusion guidance.
\newblock \emph{arXiv preprint arXiv:2207.12598}, 2022.

\bibitem[Hong et~al.(2023)Hong, Ding, Zheng, Liu, and Tang]{hong2022cogvideo}
Wenyi Hong, Ming Ding, Wendi Zheng, Xinghan Liu, and Jie Tang.
\newblock Cogvideo: Large-scale pretraining for text-to-video generation via transformers.
\newblock In \emph{ICLR}, 2023.

\bibitem[Hu et~al.(2025{\natexlab{a}})Hu, Yu, Zhou, Liang, Zhou, Lin, and Lu]{hu2025hunyuancustom}
Teng Hu, Zhentao Yu, Zhengguang Zhou, Sen Liang, Yuan Zhou, Qin Lin, and Qinglin Lu.
\newblock Hunyuancustom: A multimodal-driven architecture for customized video generation.
\newblock \emph{arXiv preprint arXiv:2505.04512}, 2025{\natexlab{a}}.

\bibitem[Hu et~al.(2025{\natexlab{b}})Hu, Yu, Zhou, Zhang, Zhou, Lu, and Yi]{hu2025polyvivid}
Teng Hu, Zhentao Yu, Zhengguang Zhou, Jiangning Zhang, Yuan Zhou, Qinglin Lu, and Ran Yi.
\newblock Polyvivid: Vivid multi-subject video generation with cross-modal interaction and enhancement.
\newblock In \emph{NeurIPS}, 2025{\natexlab{b}}.

\bibitem[Jiang et~al.(2025)Jiang, Han, Mao, Zhang, Pan, and Liu]{jiang2025vace}
Zeyinzi Jiang, Zhen Han, Chaojie Mao, Jingfeng Zhang, Yulin Pan, and Yu~Liu.
\newblock Vace: All-in-one video creation and editing.
\newblock In \emph{ICCV}, 2025.

\bibitem[Kingma and Welling(2013)]{kingma2013auto}
Diederik~P Kingma and Max Welling.
\newblock Auto-encoding variational bayes.
\newblock \emph{arXiv preprint arXiv:1312.6114}, 2013.

\bibitem[Kong et~al.(2024)Kong, Tian, Zhang, Min, Dai, Zhou, Xiong, Li, Wu, Zhang, et~al.]{kong2024hunyuanvideo}
Weijie Kong, Qi~Tian, Zijian Zhang, Rox Min, Zuozhuo Dai, Jin Zhou, Jiangfeng Xiong, Xin Li, Bo~Wu, Jianwei Zhang, et~al.
\newblock Hunyuanvideo: A systematic framework for large video generative models.
\newblock \emph{arXiv preprint arXiv:2412.03603}, 2024.

\bibitem[Li et~al.(2025)Li, Qian, Su, Diao, Xia, Liu, Yang, Zhang, and Yuan]{li2025bindweave}
Zhaoyang Li, Dongjun Qian, Kai Su, Qishuai Diao, Xiangyang Xia, Chang Liu, Wenfei Yang, Tianzhu Zhang, and Zehuan Yuan.
\newblock Bindweave: Subject-consistent video generation via cross-modal integration.
\newblock \emph{arXiv preprint arXiv:2510.00438}, 2025.

\bibitem[Lipman et~al.(2022)Lipman, Chen, Ben-Hamu, Nickel, and Le]{lipman2022flow}
Yaron Lipman, Ricky~TQ Chen, Heli Ben-Hamu, Maximilian Nickel, and Matt Le.
\newblock Flow matching for generative modeling.
\newblock \emph{arXiv preprint arXiv:2210.02747}, 2022.

\bibitem[Liu et~al.(2023)Liu, Li, Wu, and Lee]{liu2023visual}
Haotian Liu, Chunyuan Li, Qingyang Wu, and Yong~Jae Lee.
\newblock Visual instruction tuning.
\newblock \emph{NeurIPS}, 2023.

\bibitem[Liu et~al.(2024)Liu, Liu, Zhou, Xu, Xie, Han, P{\'e}rez, Liu, Kahatapitiya, Jia, et~al.]{liu2024mardini}
Haozhe Liu, Shikun Liu, Zijian Zhou, Mengmeng Xu, Yanping Xie, Xiao Han, Juan~C P{\'e}rez, Ding Liu, Kumara Kahatapitiya, Menglin Jia, et~al.
\newblock Mardini: Masked autoregressive diffusion for video generation at scale.
\newblock \emph{arXiv preprint arXiv:2410.20280}, 2024.

\bibitem[Liu et~al.(2025)Liu, Ma, Li, Chen, Liu, Li, Zhou, He, and Wu]{liu2025phantom}
Lijie Liu, Tianxiang Ma, Bingchuan Li, Zhuowei Chen, Jiawei Liu, Gen Li, Siyu Zhou, Qian He, and Xinglong Wu.
\newblock Phantom: Subject-consistent video generation via cross-modal alignment.
\newblock In \emph{ICCV}, 2025.

\bibitem[Peebles and Xie(2023)]{peebles2023scalable}
William Peebles and Saining Xie.
\newblock Scalable diffusion models with transformers.
\newblock In \emph{ICCV}, 2023.

\bibitem[Podell et~al.(2023)Podell, English, Lacey, Blattmann, Dockhorn, M{\"u}ller, Penna, and Rombach]{podell2023sdxl}
Dustin Podell, Zion English, Kyle Lacey, Andreas Blattmann, Tim Dockhorn, Jonas M{\"u}ller, Joe Penna, and Robin Rombach.
\newblock Sdxl: Improving latent diffusion models for high-resolution image synthesis.
\newblock \emph{arXiv preprint arXiv:2307.01952}, 2023.

\bibitem[Rahman et~al.(2023)Rahman, Lee, Ren, Tulyakov, Mahajan, and Sigal]{rahman2023make}
Tanzila Rahman, Hsin-Ying Lee, Jian Ren, Sergey Tulyakov, Shweta Mahajan, and Leonid Sigal.
\newblock Make-a-story: Visual memory conditioned consistent story generation.
\newblock In \emph{CVPR}, 2023.

\bibitem[Ren et~al.(2024)Ren, Liu, Zeng, Lin, Li, Cao, Chen, Huang, Chen, Yan, et~al.]{ren2024grounded}
Tianhe Ren, Shilong Liu, Ailing Zeng, Jing Lin, Kunchang Li, He~Cao, Jiayu Chen, Xinyu Huang, Yukang Chen, Feng Yan, et~al.
\newblock Grounded sam: Assembling open-world models for diverse visual tasks.
\newblock \emph{arXiv preprint arXiv:2401.14159}, 2024.

\bibitem[Rombach et~al.(2022)Rombach, Blattmann, Lorenz, Esser, and Ommer]{rombach2022high}
Robin Rombach, Andreas Blattmann, Dominik Lorenz, Patrick Esser, and Bj{\"o}rn Ommer.
\newblock High-resolution image synthesis with latent diffusion models.
\newblock In \emph{CVPR}, 2022.

\bibitem[Shen et~al.(2025)Shen, Jiang, Zhu, Ge, Cao, and Zheng]{shen2025identity}
Liao Shen, Wentao Jiang, Yiran Zhu, Tiezheng Ge, Zhiguo Cao, and Bo~Zheng.
\newblock Identity-preserving image-to-video generation via reward-guided optimization.
\newblock \emph{arXiv preprint arXiv:2510.14255}, 2025.

\bibitem[Team(2025{\natexlab{a}})]{kling}
Kling Team.
\newblock Kling1.6 elements to video.
\newblock \url{https://app.klingai.com/global/image-to-video/multi-id/new}, 2025{\natexlab{a}}.

\bibitem[Team(2025{\natexlab{b}})]{pika}
Pika Team.
\newblock Pika2.1 consistent character video.
\newblock \url{https://pollo.ai/consistent-character-video}, 2025{\natexlab{b}}.

\bibitem[Team()]{sstk}
ShutterStock Team.
\newblock Shutterstock video dataset.
\newblock \url{https://www.shutterstock.com}.

\bibitem[Team(2025{\natexlab{c}})]{vidu}
Vidu Team.
\newblock Vidu2.0 reference to video.
\newblock \url{https://www.vidu.com/ai-reference-to-video}, 2025{\natexlab{c}}.

\bibitem[Vaswani et~al.(2017)Vaswani, Shazeer, Parmar, Uszkoreit, Jones, Gomez, Kaiser, and Polosukhin]{vaswani2017attention}
Ashish Vaswani, Noam Shazeer, Niki Parmar, Jakob Uszkoreit, Llion Jones, Aidan~N Gomez, {\L}ukasz Kaiser, and Illia Polosukhin.
\newblock Attention is all you need.
\newblock In \emph{NeurIPS}, 2017.

\bibitem[Wan et~al.(2025)Wan, Wang, Ai, Wen, Mao, Xie, Chen, Yu, Zhao, Yang, et~al.]{wan2025wan}
Team Wan, Ang Wang, Baole Ai, Bin Wen, Chaojie Mao, Chen-Wei Xie, Di~Chen, Feiwu Yu, Haiming Zhao, Jianxiao Yang, et~al.
\newblock Wan: Open and advanced large-scale video generative models.
\newblock \emph{arXiv preprint arXiv:2503.20314}, 2025.

\bibitem[Wang et~al.(2025)Wang, Shi, Ou, Chen, Lin, Wang, Jiang, Yang, Zheng, Tao, et~al.]{wang2025koala}
Qiuheng Wang, Yukai Shi, Jiarong Ou, Rui Chen, Ke~Lin, Jiahao Wang, Boyuan Jiang, Haotian Yang, Mingwu Zheng, Xin Tao, et~al.
\newblock Koala-36m: A large-scale video dataset improving consistency between fine-grained conditions and video content.
\newblock In \emph{CVPR}, 2025.

\bibitem[Wang et~al.(2024)Wang, Wang, Tsutsui, Lin, Wen, and Kot]{wang2024evolving}
Xiyu Wang, Yufei Wang, Satoshi Tsutsui, Weisi Lin, Bihan Wen, and Alex Kot.
\newblock Evolving storytelling: benchmarks and methods for new character customization with diffusion models.
\newblock In \emph{ACM MM}, 2024.

\bibitem[Xue et~al.(2025)Xue, Yan, Wang, Liu, and Li]{xue2025stand}
Bowen Xue, Qixin Yan, Wenjing Wang, Hao Liu, and Chen Li.
\newblock Stand-in: A lightweight and plug-and-play identity control for video generation.
\newblock \emph{arXiv preprint arXiv:2508.07901}, 2025.

\bibitem[Yang et~al.(2025)Yang, Teng, Zheng, Ding, Huang, Xu, Yang, Hong, Zhang, Feng, et~al.]{yang2024cogvideox}
Zhuoyi Yang, Jiayan Teng, Wendi Zheng, Ming Ding, Shiyu Huang, Jiazheng Xu, Yuanming Yang, Wenyi Hong, Xiaohan Zhang, Guanyu Feng, et~al.
\newblock Cogvideox: Text-to-video diffusion models with an expert transformer.
\newblock In \emph{ICLR}, 2025.

\bibitem[Yuan et~al.(2025{\natexlab{a}})Yuan, He, Deng, Ye, Huang, Lin, Luo, and Yuan]{yuan2025opens2v}
Shenghai Yuan, Xianyi He, Yufan Deng, Yang Ye, Jinfa Huang, Bin Lin, Jiebo Luo, and Li~Yuan.
\newblock Opens2v-nexus: A detailed benchmark and million-scale dataset for subject-to-video generation.
\newblock \emph{arXiv preprint arXiv:2505.20292}, 2025{\natexlab{a}}.

\bibitem[Yuan et~al.(2025{\natexlab{b}})Yuan, Huang, He, Ge, Shi, Chen, Luo, and Yuan]{yuan2025identity}
Shenghai Yuan, Jinfa Huang, Xianyi He, Yunyang Ge, Yujun Shi, Liuhan Chen, Jiebo Luo, and Li~Yuan.
\newblock Identity-preserving text-to-video generation by frequency decomposition.
\newblock In \emph{CVPR}, 2025{\natexlab{b}}.

\bibitem[Zhang et~al.(2025)Zhang, Yang, Zhang, Hu, Zhu, Lin, Mei, Jiang, Peng, and Yuan]{zhang2025waver}
Yifu Zhang, Hao Yang, Yuqi Zhang, Yifei Hu, Fengda Zhu, Chuang Lin, Xiaofeng Mei, Yi~Jiang, Bingyue Peng, and Zehuan Yuan.
\newblock Waver: Wave your way to lifelike video generation.
\newblock \emph{arXiv preprint arXiv:2508.15761}, 2025.

\bibitem[Zheng et~al.(2024)Zheng, Gao, Fan, Liu, Laaksonen, Ouyang, and Sebe]{zheng2024bilateral}
Peng Zheng, Dehong Gao, Deng-Ping Fan, Li~Liu, Jorma Laaksonen, Wanli Ouyang, and Nicu Sebe.
\newblock Bilateral reference for high-resolution dichotomous image segmentation.
\newblock \emph{arXiv preprint arXiv:2401.03407}, 2024.

\bibitem[Zhou et~al.(2024{\natexlab{a}})Zhou, Zhang, Gu, Zhao, Shi, and Sun]{zhou2024sugar}
Yufan Zhou, Ruiyi Zhang, Jiuxiang Gu, Nanxuan Zhao, Jing Shi, and Tong Sun.
\newblock Sugar: Subject-driven video customization in a zero-shot manner.
\newblock \emph{arXiv preprint arXiv:2412.10533}, 2024{\natexlab{a}}.

\bibitem[Zhou et~al.(2024{\natexlab{b}})Zhou, Zhou, Cheng, Feng, and Hou]{zhou2024storydiffusion}
Yupeng Zhou, Daquan Zhou, Ming-Ming Cheng, Jiashi Feng, and Qibin Hou.
\newblock Storydiffusion: Consistent self-attention for long-range image and video generation.
\newblock In \emph{NeurIPS}, 2024{\natexlab{b}}.

\end{thebibliography}


\end{document}